\documentclass{article}

    \usepackage[preprint]{neurips_2024}

\usepackage[utf8]{inputenc} 
\usepackage[T1]{fontenc}    
\usepackage{hyperref}       
\usepackage{url}            
\usepackage{booktabs}       
\usepackage{amsfonts}       
\usepackage{nicefrac}       
\usepackage{microtype}      
\usepackage{xcolor}         
\usepackage{graphicx}
\usepackage{svg}
\newtheorem{definition}{Definition}[section]
\newtheorem{lemma}{Lemma}[section]
\newtheorem{example}{Example}[section]
\newtheorem{proof}{Proof}[section]

\title{Doppelgänger's Watch\\ A Split Objective Approach to Large Language Models}

\begin{document}

\author{%
  Shervin Ghasemlou\\
  Meta Reality Labs\\
  \texttt{sherving@meta.com} \\
  \And
  Seungwhan Moon \\
  Meta Reality Labs \\
  \texttt{shanemoon@meta.com} \\
  \And
  Aparajita Saraf \\
  Meta Gen AI \\
  \texttt{aparajitasaraf@meta.com} \\
  \And
  Ashish Katiyar \\
  Meta Reality Labs \\
  \texttt{ashishkatiyar13@meta.com} \\
  \And
  Mangesh Pujari \\
  Meta Reality Labs \\
  \texttt{mapujari@meta.com} \\
  \And
  Pinar Donmez  \\
  Meta Reality Labs \\
  \texttt{pinared@meta.com} \\
  \And
  Babak Damavandi \\
  Meta Reality Labs \\
  \texttt{babakd@meta.com} \\
  \And
  Anuj Kumar \\
  Meta Reality Labs \\
  \texttt{anujk@meta.com} \\
}

\maketitle

\begin{abstract}
In this paper, we investigate the problem of ``generation supervision'' in large language models, and present a novel bicameral architecture to separate supervision signals from their core capability, helpfulness. Doppelgänger, a new module parallel to the underlying language model, supervises the generation of each token, and learns to concurrently predict the supervision score(s) of the sequences up to and including each token. In this work, we present the theoretical findings, and leave the report on experimental results to a forthcoming publication.
\end{abstract}
\section{Introduction}
\begin{figure}[ht]
  \centering
  \includegraphics[width=1\columnwidth]{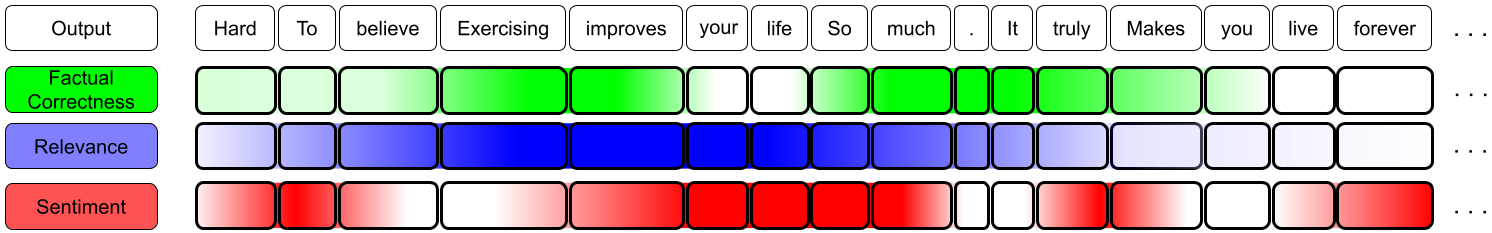}
  \caption{
  In response to ``How can exercising impact your life?'', a language model can generate a response similar to the one demonstrated here. The proposed architecture allows the language model to generate supervisory signals for the generated response up to and including each token, as the token is generated. These supervisory signals can represent objectives like factual correctness, relevance, or sentiment, which in this figure are represented by green, blue, and red colors, respectively. For simplicity, the intensity of each color indicates the score on these respective axes.}
  \label{fig:tknbytkn}
\end{figure}

Large Language Models (LLMs) have become indispensable in a variety of applications, where ensuring their response quality has become a paramount concern. While helpfulness is perceived as the core capability of language models, other supervisory signals like sentiment analysis~\citep{xing2024designing, zhan2024optimization}, factual correctness~\citep{gunjal2024molecular}, bias~\citep{chen2024humans}, and compliance~\citep{weidinger2023sociotechnical, shelby2023sociotechnical,Bianchi_2023} are among the objectives that language models can be evaluated for. However, achieving high performance on these objectives - which we refer to as supervision signal in the rest of the paper -  without compromising the acquired helpfulness is an emerging critical concern. The challenge lies in achieving a balance between these different objectives and helpfulness, as enhancements in one of these aspects could often lead to compromises in the other~\citep{bai2022training}.
\newline
In this paper, we propose an expansion of the Transformer architecture by~\citet{vaswani2017attention} into a bicameral one. We introduce a parallel component, termed the \emph{Doppelgänger}, which supervises the generation process of the tokens, concurrently evaluating the supervision score of input queries and generated responses. The architecture is agnostic to what supervisory objective it is trained for.
\newline
Contrary to most recent approaches~\citep{touvron2023llama, ge2023mart, openai2024gpt4, jiang2023mistral, geminiteam2024gemini}, where enhancing such supervisory signals is a fine-tuning step post the pretraining of the base model, our approach distinctly separates optimizing for these objective from helpfulness - keeping the LLM's helpfulness intact. This separation, as demonstrated through a mathematical proof in Section~\ref{sec:theory} offers several advantages over previous methods.
Our approach does not necessitate large-scale training as the language component of this architecture remains static during Doppelgänger's training -- hence providing the Doppelgänger with its understanding of the language. Additionally, the Doppelgänger component predicts supervision scores at the same time as the language component generates tokens(see Figure\ref{fig:tknbytkn}). This results in simpler integration and reduced latencies, particularly in applications where this capability eliminates the need for a separate model to check the response's quality. Furthermore, the Doppelgänger is agnostic to the underlying modality. In scenarios where the language model accepts inputs from other modalities~\citep{geminiteam2024gemini, moon2023anymal}, the Doppelgänger can be trained using multi-modal data without necessitating architectural changes, as it only is connected to and requires inputs from the language component.

\section{Previous work}
\label{sec:lit}
Responses generated by an LLMs unchecked for supervisory signals can have significant negative impact on their overall performance\citep{feng2023pretraining,BARMAN2024100545,qu2023unsafe,li2023privacy, shokri2017membership}. The extension to multiple modalities, such as vision, audio, and inertial modalities increases such impact due to potential for mismatches between modalities \citep{gou2024eyes, moon2023anymal}.
Setting methods like self evaluation~\citep{ren2023selfevaluation}, adding external classifiers \citep{kim2023robust, moon2023anymal}, and prompt optimization~\citep{wu2024universal, diao2023blackbox} aside, improving other supervisory signals usually is done in a similar fashion to fine tuning for helpfulness. Among these approaches are supervised instruction fine-tuning(SFT)~\citep{ouyang2022training}, reinforcement learning from human feedback (RLHF)~\citep{dai2023safe}. While both methods have shown promise, these methods do not address the inverse relation between the objectives of helpfulness and other supervisory signals as reported by~\citet{bai2022training}.

Modifications to the architectures of language models with the aim of improving different objectives independently, is an understudied field. LoRA \citep{hu2021lora} allows fine-tuning without retraining the entire model, however, training LoRA for other objective changes the behavior of the model, resulting in deteriorated helpfulness. Adding a classification head to the model \citep{arora2022director} helps with redirecting token generation to more desired outputs, but requires retraining the underlying language model for other objectives, which similarly, causes performance regression\citep{bai2022training}.
\\
To the best of our knowledge, ours is the first study to separate the optimization of supervisory signals from helpfulness, while keeping the generative capabilities of the language model untouched.
\section{Architecture}
\label{sec:arch}
In this section, we present our new bicameral architecture, which consists of the language component, for which here we use a decoder only transformer, and the Doppelgänger component, which is similar in architecture and gets input from the language component at the end of each attention module, to supervise the inputs and generated outputs.
\begin{figure}[ht]
  \centering
  \includegraphics[width=.75\columnwidth]{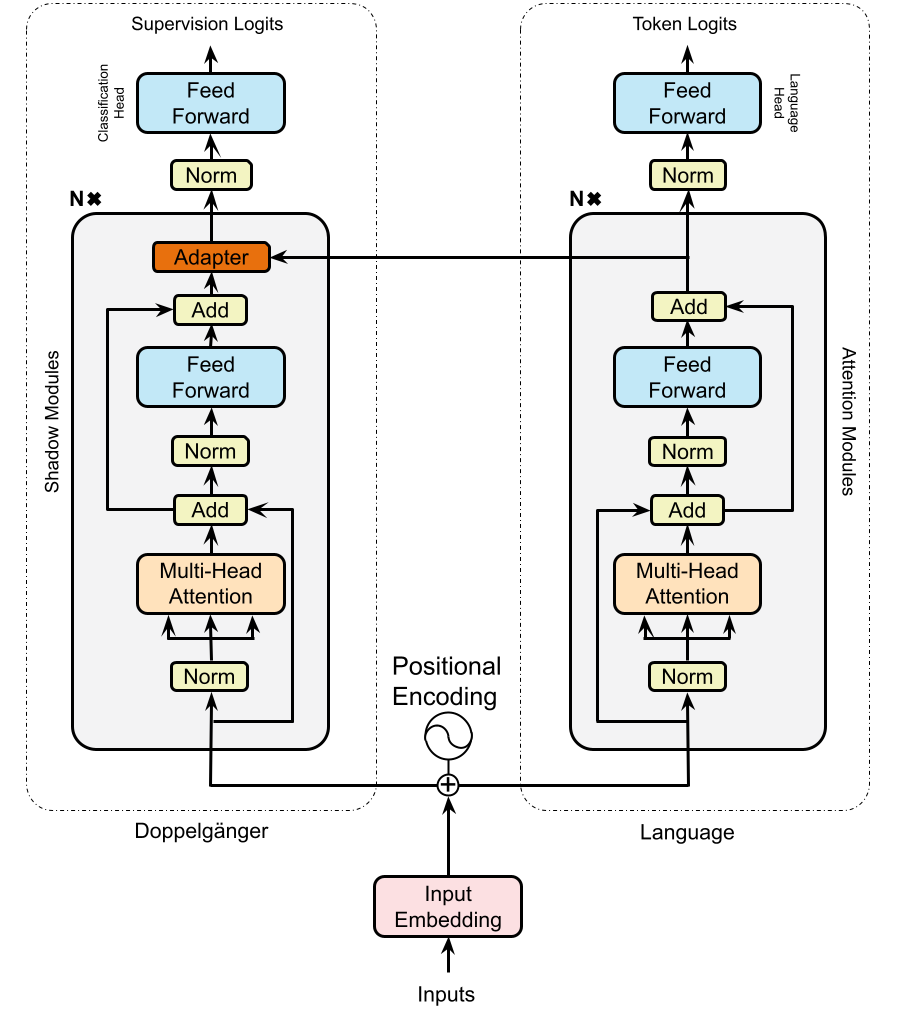}
  \caption{The proposed bicameral architecture, which is an extension to the Transformer`\citep{vaswani2017attention} architecture. This figure itself is also inspired by the illustrations provided in the same paper. Both components are decoder only transformers, consisting of $N$ attention modules, and receive the same input at their first layers, which are input embeddings enhanced by positional encoding. The output of the last attention module from the language component is used for token generation, and from the Doppelgänger component for supervision scores.}
  \label{fig:bicameral_arch}
\end{figure}
\subsection{A bicameral architecture}

As Figure~\ref{fig:bicameral_arch} Right illustrates, the architecture consists of two main components, the language component and the Doppelgänger. The attention modules are sequentially connected, and the outputs from the last layer of the language component are normalized and fed into a linear language head for sampling and token generation. Similarly, the outputs from the last layer of the supervision component(Doppelgänger) are normalized and fed into a linear classification head for score prediction. The idea is, as we explain in more details in Section~\ref{sec:theory}, to use a pretrained model for the language component that is not fine tuned for the aspect that the Doppelgänger component is to supervise, e.g, factual correctness. We keep the language component untouched -- the language component will stay frozen during the training time. The architecture keeps the auto-regressive behaviour of the underlying language model, where each generated token is appended to the concatenation of the input and previously generated sequence of tokens, to be used for the generation of the next token.

The Doppelgänger component, is an almost identical component. For simplicity, we call the attention modules of the Doppelgänger component the \emph{shadow} modules. There is one major difference between the two components: while each module in the language component accepts input from the previous attention module only, the inputs to the shadow modules are a linear transformation of the concatenation of the outputs from the previous attention module and the outputs from the previous shadow module. The exception is the first module in both components which as input receive the embeddings of the input tokens. Since Doppelgänger probes into the language component through these connections, it will be able to utilize its language understanding in predicting the supervision scores. Depending on the supervision task, this could mean Doppelgänger can have a much smaller number of parameters by avoiding replicating the language capabilities of the language component.

The proposed architecture, allows the language component to generate outputs freely without affecting its acquired helpfulness during its pretraining. The language component does not have to concern itself with other supervisory aspects of the generated response as it is left to the Doppelgänger. The Doppelgänger, concurrently, observes the refinements of the next token prediction through the chain of attention modules, and combines that info with its own understanding of the input, optimized for a specific objective, to predict the supervision score of the input sequence up to and including the newly generated tokens. The decision to allow the generated output to go through is left to the downstream applications that would utilize the Doppelgänger's score. However, we prove in Section~\ref{sec:theory} that regardless of how the decision is made, our approach is a better choice compared to current state-of-the-art alternatives.
\section{Multi Objective Optimization in Language Models}
\label{sec:theory}
In this section, we provide a set of definitions, examples, and a lemma along its proof, to establish the theoretical foundations of the proposed architecture. 
\subsection{Definitions}

\begin{definition}[Language Function]
Let $\mathcal{T}$ be an embedding space of tokens, and $\mathcal{S}$ a Cartesian product of a set of $n$ Cartesian spaces $S_i$ such that $\mathcal{S} = \prod_{i=1}^{n}S_{i}$. We define a Language function $\mathcal{L}$ as:

\[\mathcal{L}: \mathcal{T^*} \rightarrow \mathcal{S}\]

\noindent such that for any given input sequence $ (t_1, t_2, \ldots, t_m) \in \mathcal{T^*} $, where $m$ is the length of the sequence, the output is  a set $\{s_i | s_i \in S_i, 0<i\leq~n\}$.
\end{definition}

\begin{definition}[Extended Language Function]
We define a language function $\mathcal{L}: \mathcal{T^*} \rightarrow \mathcal{S}$ with $n$ outputs as an extended language function, denoted by $\mathcal{L}_{\theta_1, \theta_2, \dots, \theta_n}$, where each $\theta_i, 0 < i \leq n $ is a set of independent parameters that solely are used in determining the value of $s_i \in \mathcal{S}$.
\end{definition}
\noindent Let's provide a few examples.
\begin{example}
Most Language models are sequence to sequence models, where $n=1$ and $\mathcal{S} =\mathcal{T^*}$, where the model takes in sequences, and outputs sequences. 
\end{example}

\begin{example}
For the bicameral architecture introduced in section~\ref{sec:arch}, we have $n=2$ and $\mathcal{S} = \mathcal{T^*}\times \mathbb{R}$, where the model takes in a sequence, and outputs both another sequence, and a score. 

\end{example}

\noindent In this example, the parameters in the language head only contribute to the generation of the output sequence, and the ones in the Doppelgänger solely contribute to predicting the score. The rest of the parameters are common between the two. 
\noindent Now let's define reward functions.
\begin{definition}[Reward Function]
\raggedright A reward function $\mathcal{R}_\mathcal{S}$, maps any value from the range of a language function to a real number $r$:
\[\mathcal{R}_\mathcal{S}: \mathcal{S} \rightarrow \mathbb{R}\]
\end{definition}
\noindent 
\begin{definition}[Composite Reward Function]
A function that is the result of mapping some input reward functions $\mathcal{R}_1, \mathcal{R}_2, \dots \mathcal{R}_k$, under some map $\mathcal{M}$ into a single reward function is called a Composite Rewards Function, denoted by $\mathcal{CR}$:
\[\mathcal{CR}= \mathcal{M}(\mathcal{R}_1, \mathcal{R}_2, \dots \mathcal{R}_k),~\mathcal{M}: \mathcal{R^*} \rightarrow \mathcal{R}\]
\end{definition}

\noindent Note that this definition poses no restriction on how the reward functions are composed.


\noindent Using the definitions above, now we can present the following lemma. 
\begin{lemma}[Split Objective Supremacy]
\label{lem:lemma}
For any given extended language function $\mathcal{L}_\theta: \mathcal{T^*} \rightarrow \mathcal{S}$ and a monotonic composite reward function $\mathcal{CR}$ composed of monotonic reward functions $\mathcal{R}_{\mathcal{S}_i}: \mathcal{S}_i \rightarrow \mathbb{R}, 0 < i \leq n $, there exists another extended language function $\mathcal{L}_{\theta_1, \theta_2, \dots, \theta_n}: \mathcal{T^*} \rightarrow \mathcal{S}_1 \times \mathcal{S}_2 \times \dots \times \mathcal{S}_n$ such that $\mathcal{S} \subseteq \mathcal{S}_1 \times \mathcal{S}_2 \times \dots \times \mathcal{S}_n$, and for any given input $t \in \mathcal{T^*}$, we have 

\[\mathcal{CR}(\mathcal{L}_\theta(t)) \leq \mathcal{CR}(\mathcal{L}_{\theta_1, \theta_2, \dots, \theta_n}(t))\]
\end{lemma}
The proof is provided in Appendix~\ref{apdx:thr_sub}. Informally put, for multi objective language models in any domain, the ones with split outputs that are separately optimized for each reward function, are always at least as good as the single output ones optimized for the composite reward function. Further discussions appear in the appendix Section~\ref{apdx:thr}
\section{Conclusion and future work}
\label{sec:con}
This research introduces an innovative bicameral architecture for large language models designed to oversee the generation process while maintaining helpfulness. The proposed method incorporates a parallel component named Doppelgänger, which is designed to monitor the generation process and assess it for some supervisory score. Theoretical results suggest that this approach is expected to perform at least as well as, if not better than, existing methods currently in use in the field. Detailed experimental results will be presented in a forthcoming paper.

\clearpage

\clearpage
\appendix

\section{Appendix}
\label{apdx:thr}

\subsection{Split Objective Supremacy: Proof and Theoretical Implications}
\label{apdx:thr_sub}
In this section we first provide a proof for the lemma presented in Section~\ref{sec:theory}. We also further explain and provide context for this lemma and the ideas shaping it.
\begin{proof}
Given $s = \mathcal{L}_\theta(t)$ we have

\small \[\mathcal{CR}(\mathcal{L}_\theta(t)) = \mathcal{CR}(\mathcal{R}_{\mathcal{S}_1}(s_{S_1}), \mathcal{R}_{\mathcal{S}_2}(s_{S_2}), \ldots, \mathcal{R}_{\mathcal{S}_n}(s_{S_n}))\]
\normalsize	
\noindent where $s_{S_i}$ is the image of $s$ on $S_i$. Assume $\theta$ is in an optimum state with respect to $\mathcal{CR}$, which implies $\mathcal{CR}$ is the Pareto front for all of its composing reward functions $\mathcal{R}_{\mathcal{S}_i}(s_{S_i})$ -- so no $\mathcal{R}_{\mathcal{S}_i}(s_{S_i})$ can be optimized further without making at least one other $\mathcal{R}_{\mathcal{S}_j}(s_{S_j}), j \neq i$ worse. Given that, another extended language function $\mathcal{L}_{\theta_1, \theta_2, \dots, \theta_n}: \mathcal{T^*} \rightarrow \mathcal{S}_1 \times \mathcal{S}_2 \times \dots \times \mathcal{S}_n$ can be built where each $\theta_i$ is optimized with respect to $\mathcal{R}_{\mathcal{S}_i}$ independently, where an optimum state of $\theta_i$ is at least as optimum as the state of $\theta$ with respect to $\mathcal{R}_{\mathcal{S}_i}$. They can be equally optimum when either $\theta_i=\theta$, or, when $\theta$ not only is optimum with respect to $\mathcal{CR}$, but also it is optimum with respect to $\mathcal{R}_{\mathcal{S}_i}$.  Therefor, for each $\mathcal{R}_{\mathcal{S}_i}, 0 < i \leq n $, given $(s_1, s_2, \dots, s_n) = \mathcal{L}_{\theta_1, \theta_2, \dots, \theta_n}(t)$, we have:

\[\mathcal{R}_{\mathcal{S}_i}(s_{S_i}) \leq \mathcal{R}_{\mathcal{S}_i}(s_i)\]

\noindent Consequently, since $\mathcal{CR}$ is monotonic with respect to the composing reward functions, we can infer: 
\[\mathcal{CR}(\mathcal{L}_\theta(t)) \leq \mathcal{CR}(\mathcal{L}_{\theta_1, \theta_2, \dots, \theta_n}(t))\]

\noindent If $\theta$ is not in an optimum state with respect to $\mathcal{CR}$, then, a fortiori, the argument holds. 
\end{proof}

A common practice for optimizing LLMs for different objectives has been to assign each a reward function which could involve human evaluators. The language model usually has only one kind of output generated, a sequence of tokens, where any metric, including helpfulness, should be inferred from that output -- which in Lemma~\ref{lem:lemma} we denote such single outputs by $s$. The projections of it on each $S_i$ subspace for each objective, is $s_{S_i}$. The model is optimized with respect to some combination of the reward functions, which we model by composite reward functions, and can take place through pretraining, fine-tuning, or other approaches described in Section~\ref{sec:lit}. The composite reward function described in Section~\ref{sec:theory} is defined such that it is agnostic to how the reward functions are combined or in what order applied. As long as the composite reward function is monotonic in respect to the composing reward functions the lemma holds. The monotonicity assumption, is not a limiting one as in real world applications, usually better helpfulness, better factual correctness or better other metrics, mean a better language model.

Lemma~\ref{lem:lemma} states that we can almost always build a model that has a different set of parameters, each yielding a separate output, per each objective(i.e., reward function) that would yield a higher composite reward in comparison to single output alternatives, using the exact same reward functions and composite reward function. The only exception, mathematically speaking, is when the objective functions are all separable and the set of parameters in the single output model, $\theta$,  is in an optimum state for all the objective functions, in which case the two language models would perform the same with respect to given composite reward functions. In such a model, all parameter sets $\theta_i$ should be independent from the others, and should be optimized with respect to their corresponding reward functions. 

While the bicameral architecture presented in Section~\ref{sec:arch} is designed based on the same idea, it should be pointed out that this specific architecture is not necessarily the best architecture that can achieve split objective supremacy.



\subsection{Discussion}
\label{sec:disc}
This work introduces a novel bicameral architecture aimed at enhancing the supervision of language models without altering their inherent language capabilities. The core idea is to replace the traditional fine-tuning process with a supervision mechanism that operates in parallel to the generative component. This allows the model to predict supervision scores for both inputs and generated tokens concurrently during generation.

The current work is also centered around the idea that large language models, pre-trained on large corpus of data are inherently helpful. The architecture leverages this idea to preserve the helpfulness and avoid the adverse effects reported in other studies~\cite{bai2022training}, by introducing the Doppelgänger component. This component is designed to be trained to supervise the original, "untouched" language model, enabling it to generate content freely while still being under supervision. The supervision can be learned through labeled training data, which helps the Doppelgänger component to effectively score or classify both inputs and outputs. The probes from the language component to the Doppelgänger, also enable it to predict or classify at the same time as token generation. Without such probes, the Doppelgänger would be able to predict only up to the token before the one that is being generated, and therefor, always a token behind. 
In contrast to approaches like generator classifiers\citep{arora2022director}, where the language component remains frozen and only late fusion classification heads are used, our bicameral approach does not compromise the language capabilities. Unfreezing the language component or integrating a LoRA\cite{hu2021lora} component could potentially enhance supervision but at the cost of altering the language capabilities. The proposed architecture, however, is expected learn and predict supervisory signals without modifying the language component, as supported by our mathematical results.

A significant limitation of this study is the absence of experimental results, which are currently being conducted and will be detailed in a forthcoming paper. Future work could also expand this architecture to include bidirectional interactions between the language and Doppelgänger components. This would be particularly advantageous in real-time user interactions, allowing the system to dynamically adjust and redirect the responses based on supervisory scores.

\clearpage

\end{document}